\documentclass[journal]{IEEEtran}
\ifCLASSINFOpdf
\else
\fi

\usepackage{stfloats}

\makeatletter
\newcommand*{\centerfloat}{%
  \parindent \z@
  \leftskip \z@ \@plus 1fil \@minus \textwidth
  \rightskip\leftskip
  \parfillskip \z@skip}
\makeatother

\def\rot{\rotatebox}

\usepackage[utf8]{inputenc}
\usepackage{url}
\usepackage{booktabs}
\usepackage{float}
\restylefloat{table}
\usepackage{lineno,hyperref}
\modulolinenumbers[5]
\usepackage{xcolor}
\usepackage[pdftex]{graphicx}

\begin{document}
\title{LNEMLC: Label Network Embeddings for Multi-Label Classification}

\author{Piotr~Szymański,
        Tomasz~Kajdanowicz,
        and~Nitesh~V.~Chawla,~\IEEEmembership{Senior Member}% <-this % stops a space
\thanks{P. Szymański and T. Kajdanowicz are with the Department
of Computational Intelligence, Wroclaw University of Science and Technology, Poland, e-mail: \{piotr.szymanski, tomasz.kajdanowicz\}@pwr.edu.pl}% <-this % stops a space
\thanks{Nitesh V. Chawla is wit the Department of Computer Science and Engineering, University of Notre Dame, Notre Dame, USA and with the Department of Computational Intelligence, Wroclaw University of Science and Technology, Poland, e-mail: nchawla@nd.edu}% <-this % stops a space
\thanks{Manuscript received December XX, 2018; revised August 26, 2015.}}


\markboth{Journal of \LaTeX\ Class Files,~Vol.~14, No.~8, August~2015}%
{Shell \MakeLowercase{\textit{et al.}}: Bare Demo of IEEEtran.cls for IEEE Journals}



\maketitle

\begin{abstract}
Multi-label classification aims to classify instances with discrete non-exclusive labels. Most approaches on multi-label classification focus on effective adaptation or transformation of existing binary and multi-class learning approaches but fail in modelling the joint probability of labels or do not preserve generalization abilities for unseen label combinations. To address these issues we propose a new multi-label classification scheme, LNEMLC - Label Network Embedding for Multi-Label Classification, that embeds the label network and uses it to extend input space in learning and inference of any base multi-label classifier.  The approach allows capturing of labels' joint probability at low computational complexity providing results comparable to the best methods reported in the literature. We demonstrate how the method reveals statistically significant improvements over the simple kNN baseline classifier. We also provide hints for selecting the robust configuration that works satisfactory across data domains.

\end{abstract}


% \fi
\IEEEpeerreviewmaketitle

\begin{figure*}[!b]
    \centering
    \includegraphics[width=\textwidth]{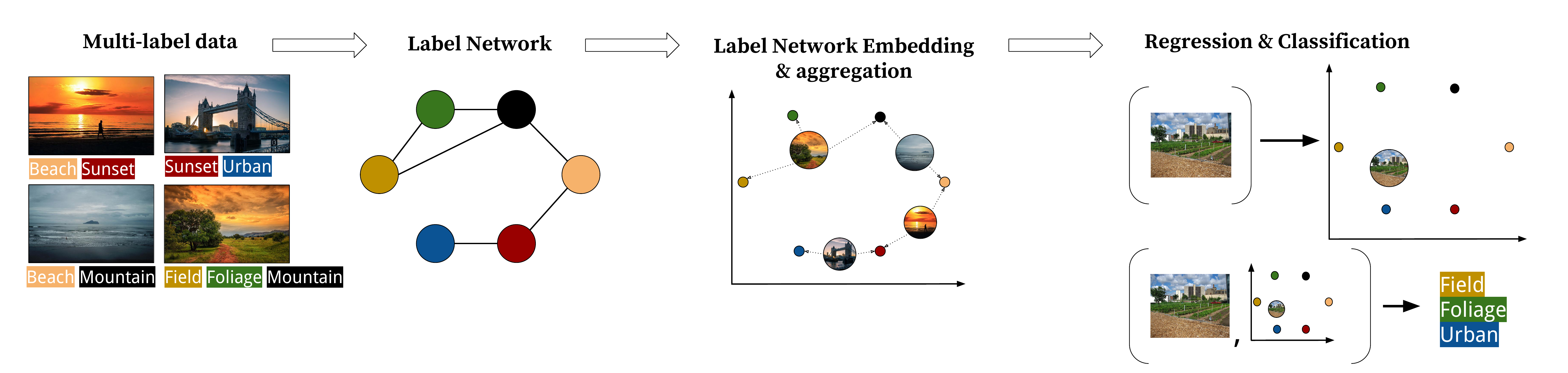}
    \caption{Label Network Embeddings for Multi-Label Classifiation scheme example diagram based on the scene data set.}
    \label{fig:my_label}
\end{figure*}
\section{Introduction}

In our daily life, we continuously encounter data classified with multiple categories or labels. Be it youtube videos, Instagram photos, articles in newspapers or more recently even our genome on gene analysis websites, we depend on possible categories or labels to assimilate and reconcile information. Such labels usually denote the simplest understandable terms, while it is from how they occur together that creates sophisticated concepts and contexts. %It emerges in the label networks. 
This problem is distinct from typical classification problems wherein each instance has one label. In this case, an instance may be categorized into multiple labels. 

However there are no low complexity multi-label methods that learn the joint label probabilities and dependencies of multi-label data, achieve high accuracy and do not require costly parameter optimization.

To fill this gap we propose LNEMLC: Label Network Embeddings for Multi-Label Classification, a new low-complexity approach to multi-label classification built on top of two intuitions that embedding a label space may improve classification quality and that label networks are a viable source of information in multi-label problems. We thus propose a novel approach for multi-label embeddings that joins the predicting power of single-label methods with discriminative power of multi-label embeddings.

LNEMLC (1) yields statistically significant improvements over a baseline classifier; (2) learns joint conditional distribution under a new inference scheme; has much (3) lower complexity than established problem transformation methods and embeddings.

We review the state of the art of multi-label classification and network embeddings in Section \ref{sec:mlc}, lay out our how LNEMLC works in Section \ref{sec:method} and describe the experimental setup in Section \ref{sec:setup}. In Section \ref{sec:results} we proceed to discuss the potential of label network embeddings, compare a variety of embedding methods and parameters and evaluate LNEMLC's performance compared to existing methods. We summarise our findings, conclusion  and ideas for future work in Section \ref{sec:conclusion}.

\section{Related work}
\label{sec:mlc}

\subsection{Multi-label classification}

Multi-label classification is a task of learning a function $f : X \rightarrow Y$ that maximizes a generalization quality measures $q (Y \times Y) \rightarrow \mathcal{R} $. Usually $f$ is called the classifier, is trained on a subset of data called the training set: $\hat{X}$, $\hat{Y}$; and is evaluated on test data $\bar{X}$, $\bar{Y}$ using the measure $q$: $q(\bar{Y}, f(\bar{X}))$. We lay out the notation used throughout this paper in Table \ref{tab:mlnotation}. Multi-label generalization quality measures can be divided into example and label-based depending on which order of relations between labels they take into account.  We describe selected measures in in Section \ref{sec:setup}.

\begin{table}[t]
    \centering
    \begin{tabular}{l|p{7cm}}
    \toprule
        Notation & Description \\
        \midrule
    $L$ & the the label set, $|L| = l$ \\
    $X$ & the multi-label input space, shaped $n \times m$ \\
    $Y = 2^L$ & the multi-label input space \\
    $E = R^{d}$ & denotes the embedded space  \\
    $V$ & the matrix of embedded label vectors of dimensions $l \times d$, $V_i$ is the embedding of the \textit{ith} label \\
    $m$ & the number of features in the problem \\ 
    $n$ & the number of instances \\
    $d$ & the number of dimensions \\
    \midrule
    $\hat{\cdot}$ & denotes a training context of a given symbol, i.e. $\hat{X} \subset X$ denotes the training samples from the input space, $\hat{Y} \subset Y$ denotes the training samples from the output space \\
    $\bar{\cdot}$ & denotes a testing context, i.e. $\bar{X} \subset X$ denotes the test samples from the input space, $\bar{Y} \subset Y$ denotes the test samples from the output space \\
    \midrule
    $\Phi$ & denotes the regression function used in different methods \\
    $\Theta$ & denotes the classification function used in label embedding scenarios \\
    $\eta$ & denotes the function generating the embedding \\
    $\xi$ & denotes the function used to aggregate the embeddings of each set of labels assigned to a sample \\
    \bottomrule
    \end{tabular}
    \caption{Notation for describing elements of multi-label problems.}
    \label{tab:mlnotation}
\end{table}

What differentiates multi-label problems from their single-label counterparts if the lack of mutual exclusiveness between the output variables or classes. In single-label classification, the model would select only one class, which made the learning the joint probability distribution the same task as learning the marginal probability distributions as classes would not occur together for one instance. Traditionally (\cite{tsoumakas2007multi}, \cite{zhang2014review}) multi-label problem solving strategies fall into four categories:

\begin{itemize}
    \item adapting single-label methods to take labels into consideration
    \item transforming the multi-label setting into a set of single-label tasks
    \item ensemble approaches that train one or more of the above approaches and perform additional steps to improve classification results
    \item embedding the label matrix in an embedding space and use regressors to predict embeddings for new samples, followed by a distance-based classifier used to classify the embedded representation
\end{itemize}

Madjarov et. al. \cite{madjarov2012extensive} compared a variety of multi-label classification methods. Their work includes results on established benchmark data sets from the MULAN (\cite{mulan}) library in several multi-label generalization quality measures. Their experimental scenario included extensive parameter search and has been the most often used source for performance comparison while introducing a new method to state of the art. We follow the same scenario in this paper and compare to these methods which performed successfully in  Madjarov's comparison. As there is no clear best performing multi-label classification method, we will compare to all of the methods present in \cite{madjarov2012extensive} that finished shed building a model on evaluated data sets.

Multi-label advances in recent years also include developments in a sub-area called extreme multi-label where new methods emerged such as deep-learning based domain-specific approaches for: image classification frameworks (\cite{zhao2015deep}, \cite{wang2016cnn}, \cite{wei2016hcp}), text (\cite{liu2017deep}, \cite{yen2017ppdsparse}). The field also includes tree-based \cite{prabhu2014fastxml} or embedding-based \cite{bhatia2015sparse} approaches. While extreme multi-label is an interesting task, it differs strongly from classical multi-label classification in performance expectations,  benchmark data sets and measures and falls beyond the scope of this paper. However, we note that LNEMLC is perfectly usable for extreme multi-label scenarios due to low complexity impact on the base classifier and the scalability of network embedding methods well beyond the sizes of label networks present in extreme multi-label sets.

\subsection{Algorithm adaptation}

Algorithm adaptations are based on modifying the decision principle of a single-label method to take into account label relations, these include:
\begin{itemize}
    \item modifying a decision function such as node impurity in tree-based methods, such as: (random forests of) multi-label C4.5 trees (RF-)ML-C4.5 \cite{clare2001knowledge} or of predictive clustering trees (RF-)PCTs \cite{todorovski2002ranking}
    \item applying post-classification inference approaches such as maximum likelihood estimation  on selected samples in similarity-based methods such as k nearest neighbours (ML-kNN \cite{zhang2007ml})
    \item one-vs-all or one-vs-rest schemas for algebraic methods such as Support Vector Machines
\end{itemize}

Algorithm adaptation methods are usually based on considering similarities or objective functions on a subset of features and learn their part of the joint label distribution. LNEMLC improves algorithm adaption methods by providing them with additional input features to discriminate on while increasing their complexity concerning the number of features by a constant. 

\subsection{Problem transformation and ensemble approaches}

\begin{table*}[t]
    \centering
    \begin{tabular}{p{3cm}c|c|c|c|c|c|c}
    \toprule
        {} & BR    & (E)CC    & CLR &     QWML &     HOMER &    RAkEL &    LNEMLC \\
    \midrule
Estimated distribution &     Marginal &    Marginal &    Pairwise &    Pairwise &    Joint &    Joint &    Joint \\
Number of base classifiers to train &    $l$ &    $n_{ens} \times l$  & $l^2$     &$l^2$ & $k$ & $k$  &    1 \\
Capable of predicting unseen label combinations & Yes & Yes & Yes &  Yes & Partial & Partially & Yes \\
\bottomrule
    \end{tabular}
    \caption{Characteristics of problem transformation and ensemble approaches compared to the proposed method. $k$ denotes hyper-parameter of a method and must be estimated externally.}
    \label{tab:char.pt}
\end{table*}

Following (\cite{dembczynski2012label}) we can divide the problem transformation and ensemble approaches based on how they view label dependency exploitation:
\begin{itemize}
 \item ignoring label dependence entirely and learning separate model per label, ex. Binary Relevance (BR) \cite{tsoumakas2007multi}
 \item exploiting the marginal probabilities $P(\lambda_{i}|\bar{x})$ to improve prediction quality for single labels. Such methods usually work based on a set of independent Binary Relevance classifiers and then correct their solutions using Bayesian inference, stacking, multivariate regression or kernel-based estimation, ex. such as Classifier Chains or their ensembles: (E)CC \cite{read2011classifier}
 \item exploiting the marginal probabilities of label pairs $P(\lambda_{i}|\bar{x})$ to improve prediction quality, such as Calibrated Label Ranking (CLR \cite{furnkranz2008multilabel}) or q-weighted voting based label ranking (QWML \cite{mencia2010efficient}).
 \item exploiting conditional dependencies $P(Y|\bar{x})$ of the joint conditional distribution $P(Y|X)$, often through transformation to one or more multi-class problems ex. Label Powerset \cite{zhang2014review} which transforms the problem to a multi-class problem where each of the label combinations is treated as separate classes, such approaches often exploit dividing the label space either by partitioning (RAkEL \cite{tsoumakas2011random}) or hierarchical divide and conquer strategies (HOMER \cite{tsoumakas_effective_2008})
\end{itemize}

LNEMLC is a meta-learning method that includes the best characteristics of different; we compare its characteristics to the above methods in Table \ref{tab:char.pt}. It is also noteworthy however that LNEMLC can also be used to improve problem transformation performance,  possibly eliminating the need for additional ensemble of classifiers due to enriching the discriminative capability of the input features by learning the embeddings representing the joint label distribution.

\subsection{Multi-label embeddings}

Multi-label embedding techniques emerged as a response to the need to cope with a large label space. First embedding approaches consider label space dimensionality reduction to improve computation time with the increase of computing power the field developed in the direction of finding representations that would express joint probability information in a more discriminative way than a binary indicator label assignment matrix. New embedding methods were developed using well-established algebraic manipulations known from statistical analysis and relying on a multi-variate regressor and a weak classifier as label assignment decision; we compare most cited methods in Table \ref{tab:emb.chars}.

\begin{table*}[t]
    \centering
    \begin{tabular}{p{6cm}|p{4.5cm}|p{1.5cm}|p{2cm}|p{2cm}}
        \toprule
        Method &    Embedding principle    & \# of regressors    & regression principle    &     classification principle \\
        \midrule
        Compressed sensing \cite{hsu2009multi}    & random projection and compression    &     $d$        & Linear        & distance minimalization  \\
        Principle Label Space Transformation \cite{tai2012multilabel}     & PCA    &     $d$        & Linear    decoding function \\
        Conditional Principal Label Space Transformation \cite{chen2012feature}    &     CCA \cite{sun2011canonical} &    $1$    & Linear & rounding \\
        Feature-aware Implicit Label Space Encoding \cite{lin2014multi}    &     matrix decomposition trough objective maximization    &     $l$        & Ridge        & rounding \\
        SLEEC \cite{bhatia2015sparse}    &     k-means partitioning and per cluster matrix decomposition via objective optimization    &     $d$        & Linear    &     k-NN \\
        CLEMS \cite{huang2017cost}        & MDS \cite{kruskal1964multidimensional} with extra cost objective optimization        & $d$ &     Random Forest        & 1-NN \\\hline
        LNEMLC    &     label network embedding    &     1    &     any    &     any \\
        \bottomrule
    \end{tabular}
    \caption{Characteristics of embedding approaches compared to the proposed method.}
    \label{tab:emb.chars}
\end{table*}

Huang and Lin (\cite{huang2017cost}) show that the best performing multi-label embedding is CLEMS. We will compare LNEMLC performance to multiples CLEMS models each trained with one of the evaluated measures as a cost function.

Most multi-label embedding methods embed the output space $X$ into the embedded space $E$, learn a regressor $\Phi : X \rightarrow E$ regressor from the input space to the embedding space and then train a classifier $\Theta : E \rightarrow Y$ to predict label assignments based on embeddings. Most current approaches use tree-based methods (usually Random Forests of CART trees) as $\Phi$. The main task of the classification phase in such a scheme is to find the location in the embedded space which represents labels most similar to the predicted embedding - the k-nearest neighbours are used as the classifiers, often 1-NNs. 

LNEMLC provides a general framework for using a regressor and label network embedding of choice to improve a base classifiers performance. Our method is constructed to allow the classifier to correct regression errors, while the complexity cost is only extending the number of features by the number of dimensions. We also use just one multi-dimensional regressor instead of learning a regressor per dimension, which caused CLEMS not to finish on some of the evaluated data sets. Our approach also does not require selecting the measure to optimize as label network embeddings provide discriminative information about the joint probability without the need to select additional cost-related optimization criteria.
 
\subsection{Network Embeddings}

Network Representation Learning (NRL), or Network Embedding (NE), aims to learn a latent representation of nodes that encapsulates the network's characteristics. As networks depict complex phenomena and in effect representation learning approaches differ based on principle, the order of relationships taken into account, the scale of network phenomena, community structure, network motifs, whether a network is a temporal, directed, bipartite or has other specific features. We select several approaches we consider relevant to label networks in multi-label classification out of methods described in current NRL surveys (\cite{2018arXiv180105852Z}, \cite{goyal2018graph}, \cite{2017arXiv171108752C}) and community-curated network embedding lists: \cite{awne} and \cite{opennePapers}. A general introduction to representational learning can be found in \cite{6472238}.

Survey authors divide NRLs into unsupervised and semi-supervised. As multi-label classification lacks additional information about label meta-structures, we only consider unsupervised NRLs. Network embedding methods differ the scale of network phenomena consider: micro-, meso- and macroscopic. The microscopic methods capture local proximity in the network; mesoscopic approaches consider structural similarity and intra-community proximity, macroscopic scale embeddings recognize global characteristics such as scale-free or small-world properties. In label networks, only the first two groups of phenomena are usually found. Available microscopic embeddings can preserve first, first and second, or second and higher order relations. These relations can be mapped to pairwise, shorter and longer label combinations in multi-label classification. Among mesoscopic embeddings intra-scale community distances preservation is the most promising as we already know that community structures can be used to improve classification quality \cite{szymanski2016data}.

As a result we decided to select three embedding approaches in this paper:  
\begin{itemize}
    \item node2vec \cite{grover2016node2vec}: a well-established approach which preserves higher-order relations in embedding distances by using walks to sample ordered sentence-like structures where nodes take the place of words and then uses the established word2vec (\cite{mikolov2013distributed}) skip-gram model for embedding node sequences.
    \item LINE \cite{tang2015line}: a first and second order relation preserving embedding approach which optimizes an embedding that minimizes a proximity function between nodes - either first or second order, or both separately and glues each of the embeddings per vertex. Negative edge-sampling (\cite{mikolov2013distributed}) is used to speed up optimization.
    \item M-NMF \cite{wang2017community}: a matrix factorization approach which preserves community structure in embedding distances by embedding the network into smaller subspaces per community and projecting these embeddings into a common space. 
\end{itemize}

In the next section, we describe the proposed approach and how network representation learning is used for multi-label classification purposes.

%\begin{figure}
%  \includegraphics[width=\columnwidth]{embedding.pdf}
%  \caption{Generating label network embeddings phase of LNEMLC.}
%  \label{fig:embeddings}
%\end{figure}

\section{Proposed Method}
\label{sec:method}

\subsection{Intuition}

\begin{figure*}
\centering
  \includegraphics[width=0.8\textwidth]{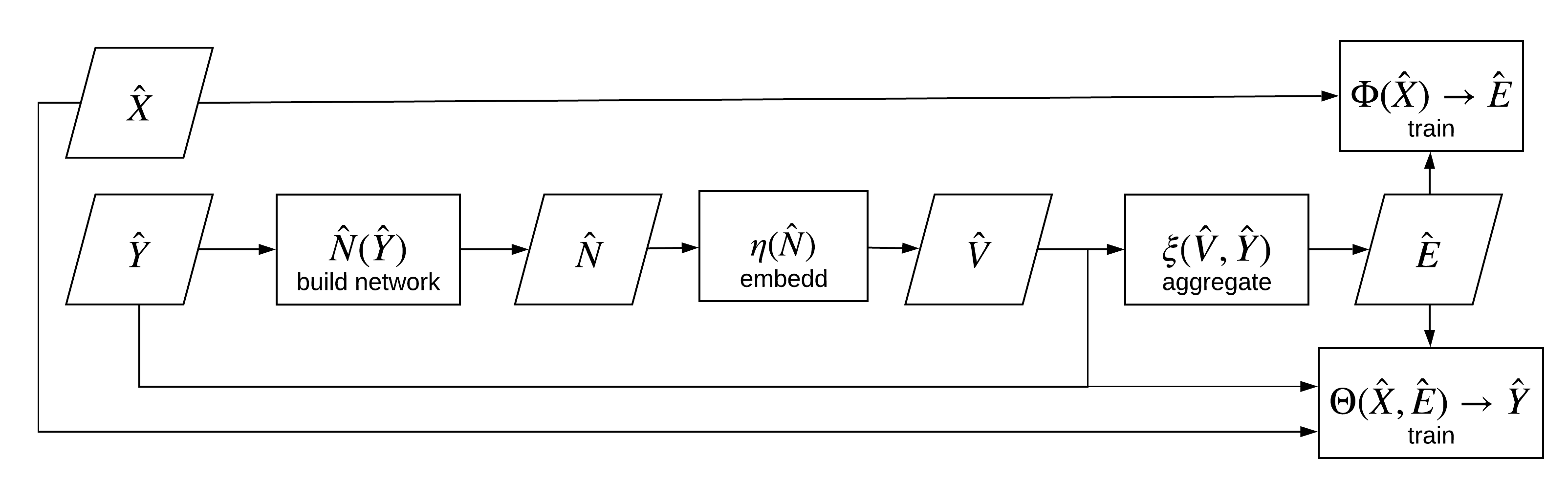}
  \caption{Graphical representation of steps in LNEMLC training phase.}
  \label{fig:training.schema}
\end{figure*}

We propose a new approach to incorporate label dependencies into multi-label classification scenarios: LNEMLC - Label Network Embedding for Multi-label Classification extends the input space visible to the classifier with an embedding of the label network constructed from the training data. 

With LNEMLC we introduce a difference in the embedding-based classification schemes as we have evaluated that it yielded better results than the original approach. Instead of using a weak 1-NN to find the example closest to our embedding, we will use a multi-label kNN that learns from both the input space and the regressed embedding to better correct the regression error present in the original scheme. Thus our classifier is $\Theta : X \times E \rightarrow Y$. This places LNEMLC between the classical regression-centered multi-label embedding setting and the traditional multi-label methods which use algorithm adaptation and problem transformation approaches. On one side multi-label embeddings should allow multi-label kNN to learn the joint probability through the possibility of discriminating over label relations, on the other the regression error from multi-label embeddings can be corrected more efficiently by taking into account both the input and embedding spaces. 

Our method builds upon the established network embedding methods that are known to have a positive impact on network classification tasks. Given an embedding method, training LNEMLC requires deciding on the number of dimensions $d$ for the embedded label vectors and on the variant of the network to embed. In this paper, we assume the network is based on label co-occurrence in both unweighted and weighted variants. This approach has yielded classification quality improvements in past work \cite{szymanski2016data}. It is nevertheless possible to deploy different methods of building label relations in the network and of weighting these relations. 

\subsection{Formal description}

The training scheme of LNEMLC consists of the following steps:
\begin{enumerate}
    \item constructing the label network, in the case of this paper, the network $\hat{N}(L,\hat{A})$ consists of node set $L$ and an edge set $A = \{(s, t): (\exists \hat{Y}_i)(\hat{Y}_{i,s} = 1 \wedge \hat{Y}_{i,t} = 1)\}$, in the weighted case 
    
    $$w(s,t) = \frac{|\{i:\hat{Y}_{i,s} = 1 \wedge \hat{Y}_{i,t} = 1)\}|}{|\hat{Y}|}$$ 
    
    \item embedding the label network $\hat{N}$ using network embedding function $\eta$ and generating a set $\hat{V}=\eta(\hat{N})$ of embedding vectors for every label assigned to at least one sample in $\hat{Y}$
    
    \item aggregating label vectors assigned to a given sample using an aggregation function $\xi$ to obtain the label space embedding for each training sample, i.e. $\hat{E}_{i} = \xi({\{v_j \in \hat{V}: \hat{Y}_{i,j} = 1\}})$ 
    
    \item training an embedding regressor $\Phi : X \rightarrow E$ on the input space $\hat{X}$ and output $\hat{E}$
    
    \item training a multi-label classifier $\Theta : X \times E \rightarrow Y$ on the input space $\hat{X} \times \hat{E}$ and output $\hat{Y}$

\end{enumerate} 

\begin{figure}[H]
  \centering
  \includegraphics[width=\columnwidth]{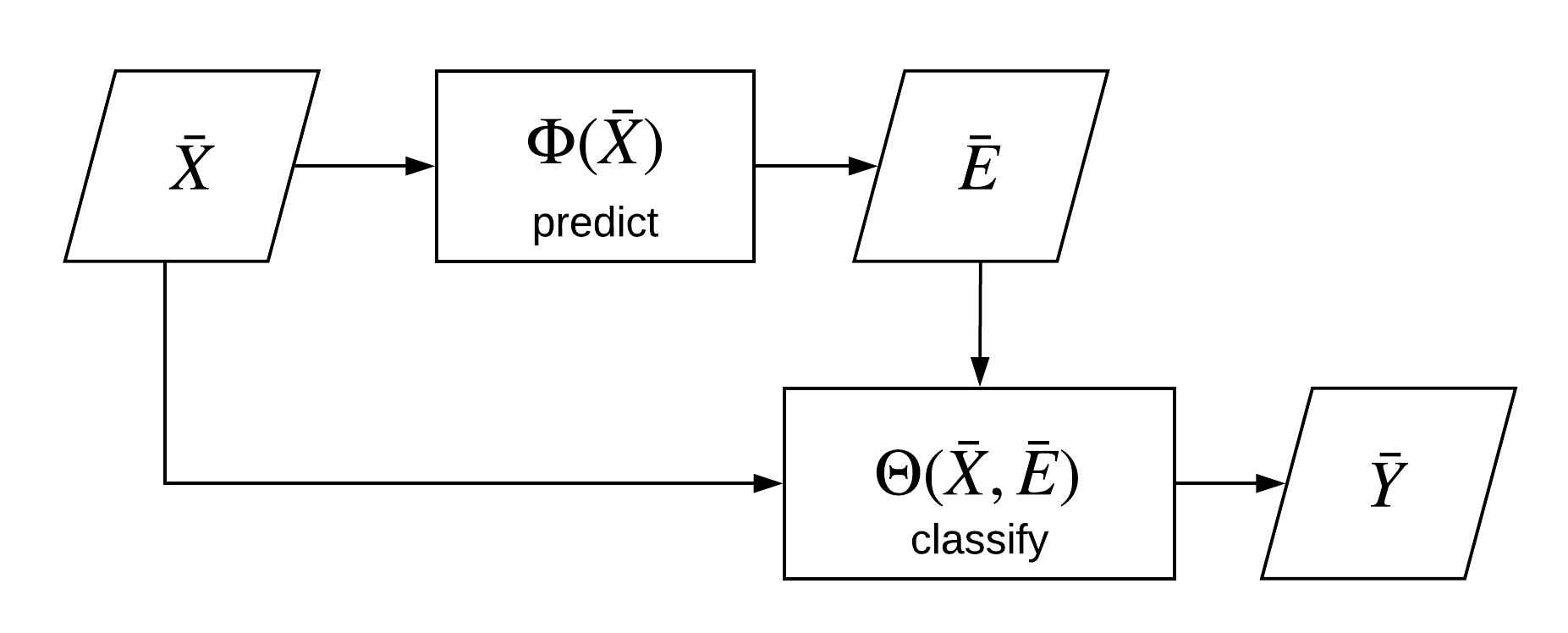}
  \caption{Graphical representation of steps in LNEMLC prediction phase.}
  \label{fig:testing.echema}
\end{figure}

The testing scheme of LNEMLC is more straightforward, the outcome of classification of the test input samples $\bar{X}$ is obtained from $\Theta(\bar{X} \times \Phi(\bar{X}))$. We provide figures that contains graphical representation of the proposed method, Figure \ref{fig:training.schema} for training and Figure \ref{fig:testing.echema} for testing.

\subsection{Complexity}

LNEMLC's complexity is the sum of the embedder, classifier and regression complexities, however our complexity is not worse than $O(ld + n{(m+d)}^2 + mn^2 + ln)$. We can see that it is dominated by kNN's quadratic complexity with the number of samples and on larger data sets the effect of using an embedding approach will have a negligible impact on the classification step, which shows a small overhead compared to a no embedding approach. 

Embedding methods have the following complexities:
\begin{itemize}
    \item LINE's complexity depends on three parameters: the negative edge sampling ratio, which is not important in the case of networks which we evaluate and we'll ignore, as it is a fixed estimated parameter equal to $5$ and from LNEMLC's perspective it is a constant; the number of edges in the network and the dimension size $d$, effectively LINE performs the embedding in $O(l_{pairs}d)$, as our network does not contain multiple edges and most multi-label problems exhibit a sparse output space, the number of labels is much closer to a constant times the number of labels and $l_{pairs} << l^2$
    \item node2vec is reported \cite{goyal2018graph} to have a complexity of $O(ld)$ although in practice there is a strong impact of random walk lengths on the execution times. In multi-label cases, however, the walk lengths are not extremely long with our parameter settings as label combinations present in real-world data are limited in length.
    \item M-NMF's complexity is $O((d+k)l^2)$ where $k$ is the size of internal representation space for detected communities, in our case $k$ is the number of communities detected by the fast greedy modularity algorithm but in general $k \leq l$ which leaves M-NMF's complexity at $O((d+l)l^2)$. In practice $k$ can also be set to a fixed dimension, the author's use a default of $20$ which would reduce the complexity to $O(d{l}^2)$
\end{itemize}

We proceed to discuss the regression and classification complexities: Linear and Ridge Regression complexity in our case amounts to $O(nm^2)$, while for CART tree based Random Forest regressor it is $O(mn\log{n})$. Nearest Neighbour methods (ML-kNN) have a training complexity of $O(mn^2 + lkn)$ and a predicting complexity of $O(mn + lk)$ where $k=5$ is a fixed constant in our experimental scenario, and usually, $k$ is fixed as obtained by parameter estimation. We can, therefore, treat Ml-kNN's complexity as $O(mn^2 + ln)$. With respect to the base classifier our method impacts only the number of features replacing the original value $m$ with $m' = m + d$, however in practical cases $d <= 4096$, and among recommended dimensions sizes $2m <= d <= 4m$ in all imaginable applications.

In the following sections, we provide experimental insights for selecting the network variant, the aggregation function $\xi$ and the dimension size $d$.

\section{Experimental setup}
\label{sec:setup}
To evaluate how well LNEMLC performs we ask of following questions:
\begin{enumerate}
    \item Do label network embeddings have the potential of improving multi-label classification?
    \item Are available regression methods capable of performing regression well enough to maintain the advantage provided by the embeddings?
    \item How does LNEMLC perform in comparison to state of the art multi-label methods?
    \item Is there a single combination of method's parameters and configuration that works satisfactorily across data domains?
    %Are there network variants, embedding space dimensions, embedding and label vector aggregation method combinations that perform better than others?
    
\end{enumerate}

We are interested in how well LNEMLC learn the joint label distribution, so we chose the subset accuracy as the quality measure. Accuracy score shows how well the method can predict exact label assignments, which is the core characteristic of a model that generalizes joint distribution well. For comparison purposes to answer questions three and four we also consider other label-based metrics from Madjarov's study. Evaluation metrics are describe in depth in \cite{madjarov2012extensive} or \cite{dembczynski2012label}.

\begin{figure}
  \includegraphics[width=\columnwidth]{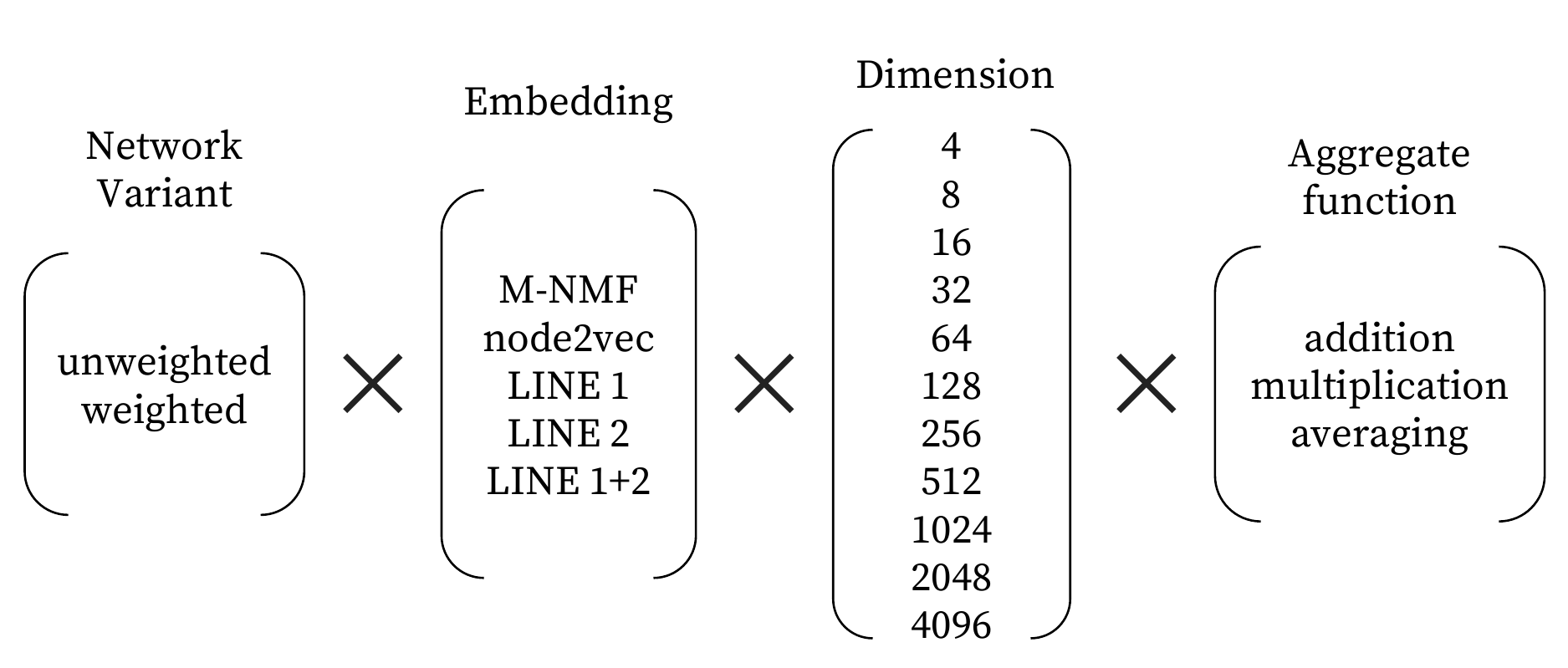}
  \caption{Combinations of embedding parameters evaluated in the 5-fold experimental scenario.}
  \label{fig:parameter.combinations}
\end{figure}

We carried out three experimental scenarios to find the answer to our questions: 
\begin{itemize}
    \item estimating the best dimensions, network variants and vector aggregation function on stratified cross-validated folds on the training data as illustrated in Figure \ref{fig:parameter.combinations}. Using the result from this scenario, we can provide a basis for the next experimental scenarios and partially answer the question no. four by providing the best performing parameter combination.
    \item the classical train/test classification with exact embeddings and regressors trained for embedding prediction of test samples for the best performing parameters per set per measure. Using this data, we answer the question no. one and two by showing the impact of exact and regressed embeddings on the base classifier generalization quality and question no. three by comparing these measures to results from Madjarov's comparison and CLEMS performance.
    \item the classical train/test classification with exact embeddings and regressors trained for the parameter set we found most likely to perform well overall. We evaluate this impact of our proposed parameter setting to finish answering question 4 with confirmation of LNEMLC's practical usefulness.
\end{itemize}

\subsection{Tools and parameters}

Label Networks based on label co-occurrence were generated using the scikit-multilearn\cite{scikit-multilearn} library. We used the following implementations for network embeddings:
\begin{itemize}
    \item node2vec: Elior Cohen's node2vec package was used\footnote{\url{https://github.com/eliorc/node2vec}}, 200 walks were performed with maximum walk length equal to twice the number of the maximum number of labels assigned to a sample in the training data
    \item LINE: the OpenNE implementation\footnote{\url{https://github.com/thunlp/OpenNE}} was used; three variants of LINE were evaluated: preserving first-order relations in the network (order: 1), second-order relations in the network (order: 2), and both (order: 1+2 in the paper, 3 in our internal parameter numbering).
    \item M-NMF: the code from the authors' GitHub repository\footnote{\url{https://github.com/benedekrozemberczki/M-NMF}} was used. NMF required a clusters parameter that denoted the number of clusters to use internally for embedding; this was set to the number of communities detected on the weighted label network based using a fast greedy modularity approach \cite{modularity} provided by the igraph \cite{igraph} package and scikit-multilearn. The M-NMF implementation crashed on several label networks; we only evaluated data sets for which NMF completed the embedding generation process. These are (number of clusters in parenthesis): bibtex (4), delicious (4), emotions (6), mediamill (3), scene (4), tmc2007\_500 (10), yeast (7).
    \item CLEMS: the code from the author's GitHub repository\footnote{\url{https://github.com/ej0cl6/csmlc}} was used with default parameters as presented in the paper. 
\end{itemize}

All network embeddings included self-edges of labels because the evaluated embeddings failed to generate an embedding vector for labels that had no edges.

Scikit-learn \cite{scikitlearn} implementation of regressors: Linear, Ridge and Random Forest are used as the function $\Phi$. The multi-label k Nearest Neighbors classifier serves as $\Theta$.  Scikit-multilearn implementation of the iterative stratification (\cite{sechidis2011stratification}, \cite{pmlrstratification}) was used. We used 5-fold cross-validation, and the iterative stratification was set to preserve equal label pair evidence distribution among folds.

We evaluated powers of two as dimension values, from 4 to 4096, unweighted and weighted variants of label networks. The weights contained the percentage of samples in the training set that were labelled with the given two labels. We evaluated addition, multiplication and averaging as label embedding vector aggregation functions. 

\subsection{Data sets}

\begin{table}[H]
    \centering
    \resizebox{.5\textwidth}{!}{\begin{tabular}{llrrrrr}
    \toprule
    {} &      domain &  instances &  attributes &  labels &  cardinality &  distinct \\
    name        &             &            &             &         &              &           \\
    \midrule
    bibtex      &        text &       7395 &        1836 &     159 &        2.402 &      2856 \\
    corel5k     &      images &       5000 &         499 &     374 &        3.522 &      3175 \\
    delicious   &  text (web) &      16105 &         500 &     983 &       19.020 &     15806 \\
    emotions    &       music &        593 &          72 &       6 &        1.869 &        27 \\
    enron       &        text &       1702 &        1001 &      53 &        3.378 &       753 \\
    mediamill   &       video &      43907 &         120 &     101 &        4.376 &      6555 \\
    medical     &        text &        978 &        1449 &      45 &        1.245 &        94 \\
    scene       &       image &       2407 &         294 &       6 &        1.074 &        15 \\
    tmc2007\_500 &        text &      28596 &       49060 &      22 &        2.158 &      1341 \\
    yeast       &     biology &       2417 &         103 &      14 &        4.237 &       198 \\
    \bottomrule
    \end{tabular}}
    \caption{Multi-label data set statistics.}
    \label{tab:datastats}
\end{table}

We used 10 data sets from MULAN's \cite{mulan} multi-label data set repository which provided a train/test division. These data sets span a variety of domains, ranging from a small to a large number of features, instances and labels and are well established as benchmark data in multi-label classification literature. Their statistical properties are presented in Table \ref{tab:datastats}.

Bibtex \cite{katakis_multilabel_2008} is a bag-of-words data set concerning. Corel5k (\cite{duygulu_object_2002}) and scene (\cite{boutell_learning_2004}) are image data sets with labels denoting the clipart contents. Delicious (\cite{tsoumakas_effective_2008}) is a bag-of-words representation of website contents bookmarked in the del.icio.us website with labels representing tags added to a given bookmark. Emotions (\cite{trohidis2008multi}) is an audio data set labelled with categories of emotions the Tellegen-Watson-Clark model. Enron (\cite{klimt2004enron}) contains emails from senior Enron Corporation employees categorized into topics by the UC Berkeley Enron E-mail Analysis Project\footnote{\url{http://bailando.sims.berkeley.edu/enron_email.html}} with the input space being a bag of word representation of the e-mails. Mediamill (\cite{snoek_challenge_2006}) is a set of features extracted from videos labelled with tags concerning the video content. The medical (\cite{read_classifier_2011}) dataset is a bag-of-words representation of patient symptom history and labels represent diseases following the International Classification of Diseases. Tmc2007 (\cite{tsoumakas_effective_2008}) contains an input space consisting of similarly selected top 500 words appearing in flight safety reports. The labels represent the problems being described in these reports. The \textit{yeast} \cite{elisseeff_kernel_2001} data set concerns the problem of assigning functional classes to genes of \textit{saccharomyces cerevisiae} genome.

\section{Results\label{sec:results}}

In this section, we present the results of LNEMLC, the proposed method, organized by research question. 

\subsection{Do label network embeddings have the potential of improving multi-label classification?}

To evaluate the potential of label network embedding applications to multi-label classification, we compare the results obtained by exact embeddings to no embedding baselines (N/E) for each of the embeddings, next compare the embedding approaches to themselves, methods from Madjarov's comparison and best classical multi-label embeddings. Exact embeddings are the direct embeddings of the known test data performed in the same way as the training embeddings, without the need for regression. While this scenario is impossible in practice, as it assumes that the regressor $\Phi$ would have made no mistakes, it allows us to see the upper bound of LNEMLC potential.

\begin{figure}[htb]
    \centering
    \includegraphics[width=\columnwidth]{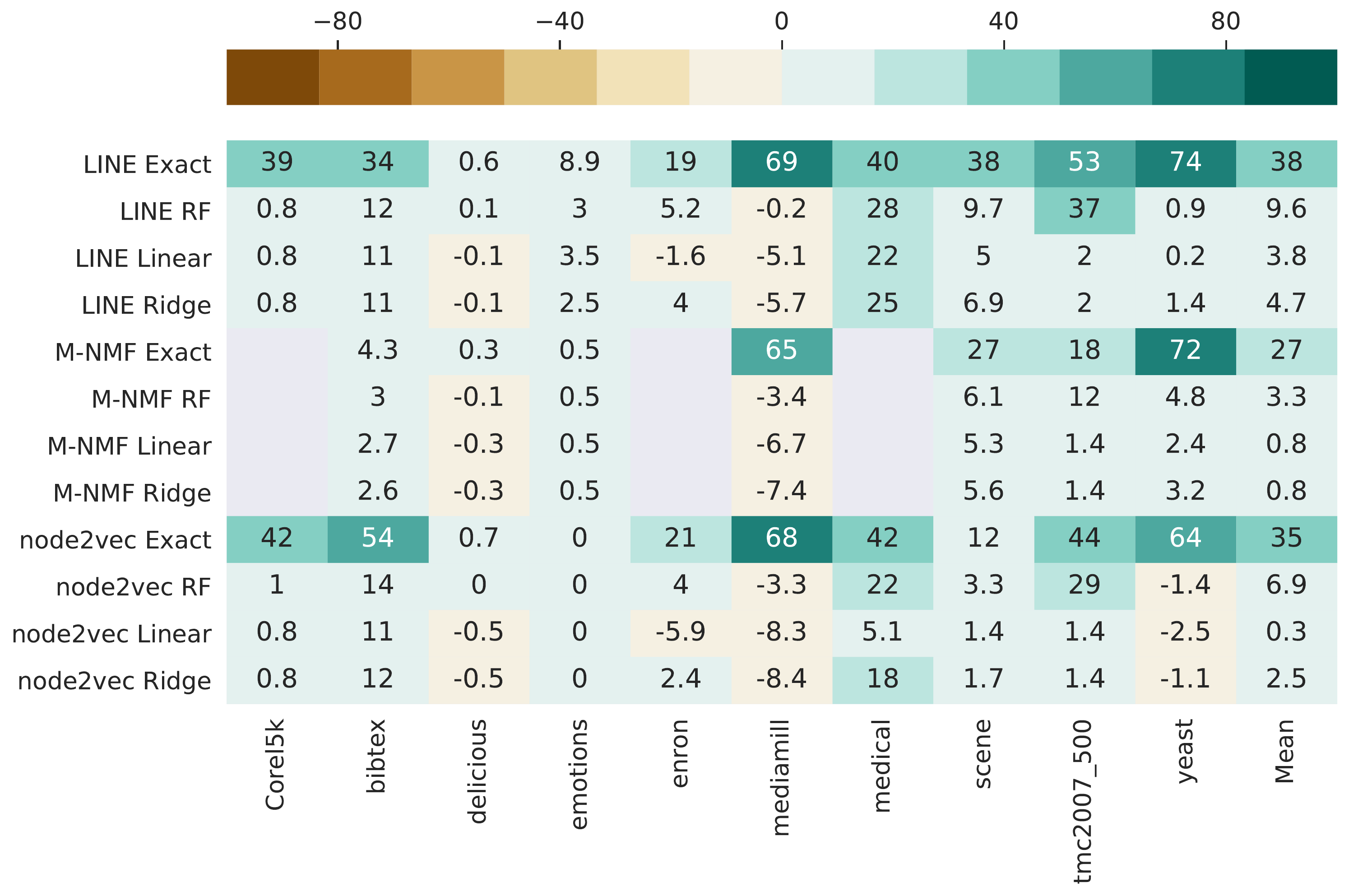}
    \caption{Performance difference between LNEMLC with exact embeddings and no-embedding baseline (LNEMLC Exact - Baseline) with respect to common multi-label metrics with LINE embedded label network information. Differences for each measure are statistically significant as compared using Wilcoxon's signed rank test. The tests p-values for LNEMLC variants are: LINE Exact: \textbf{0.0051}, LINE RF: \textbf{0.0093}, LINE Linear: 0.1394, LINE Ridge: \textbf{0.0469}, M-NMF Exact: \textbf{0.0180}, M-NMF RF: 0.1282, M-NMF Linear: 0.3105, M-NMF Ridge: 0.3105, node2vec Exact: \textbf{0.0077}, node2vec RF: 0.0797, node2vec Linear: 0.8588.}
    \label{fig:accuracy.improvements}
\end{figure}

Figure \ref{fig:accuracy.improvements} shows the differences in accuracy scores between how LNEMLC performed with exact and regressed embeddings over the baseline kNN for each of the embedding functions $\eta$. All differences are presented as percentage points. We can see that the potential of exact embedding's improvements is large, from 12 to 74 percentage points on most data sets, 27-38 percentage points on average. However, we note that for the delicious data set an improvement of 0.6 or 0.7 percentage points is a very high correction given that first three best-performing methods in accuracy in Madjarov's comparison differ by 0.3 percentage point. LNEMLC with exact embeddings displays a statistically significant potential to improve multi-label classification performance of the base classifier. This potential is impressive, but often hard to achieve in a practical setting due to regression errors. We thus answer the question positively. 

We now proceed to compare LNEMLC variants to each other in the next subsection and answer the question of how regression impacts LNEMLC performance.

\subsection{Are available regression methods capable of performing regression well enough to maintain the advantage provided by the embeddings?}

As expected we see in Figure \ref{fig:accuracy.improvements} that regression approaches maintain only a fraction of the improvement potential. LNEMLC with LINE embeddings maintained the statistical significance of improving accuracy with all regressors and is the only regressed variant to do so. It yields an improvement of nearly ten percentage points on average with Random Forest regressors and around four percentage points with more rudimentary Ridge and linear regressions. 

While we can see that LNEMLC with LINE and Random Forests performs best, we perform pairwise comparisons of LNEMLC with LINE, node2vec and the no embedding baseline. We control family-wise comparison errors using the standard Friedman-Iman-Davenport multiple comparison tests with Hochberg post hoc p-value correction for pairwise comparisons. We exclude M-NMF from this comparison as the embedding did not finish on several data sets and would lower the power of the procedure. 

\begin{figure}[htb]
    \centering
    \includegraphics[width=\columnwidth]{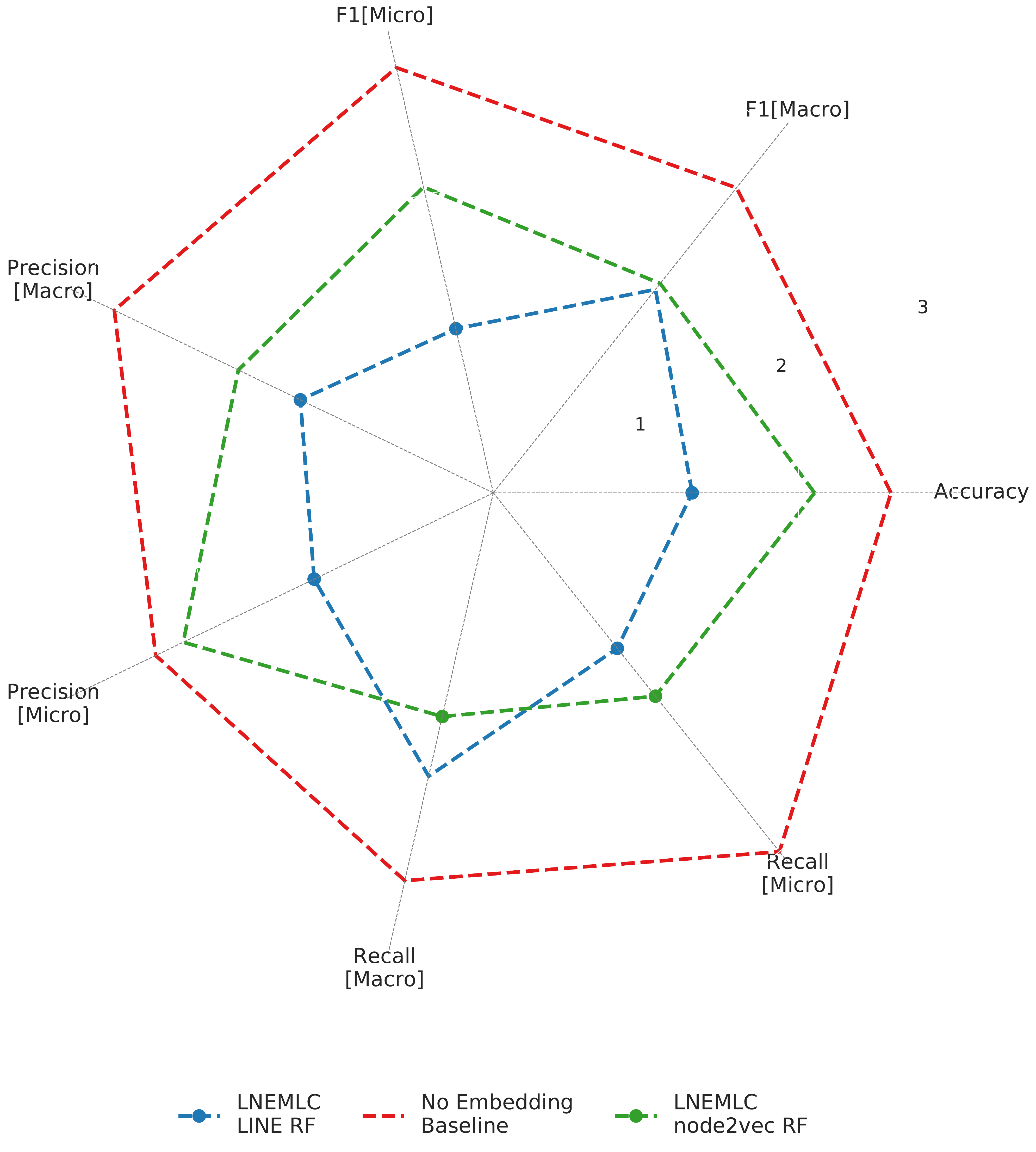}
    \caption{Mean ranks of random forest regressed LNEMLC and no embedding baseline with respect to common multi-label metrics with LINE embedded label network information.}
    \label{fig:regressor.comparison}
\end{figure}

In Figure \ref{fig:regressor.comparison} we see that LNEMLC LINE with the Random Forest regression's improvement over the base classifier without embedding is statistically significant in five out of seven measures, while it always ranks higher than the baseline. Using LNEMLC with node2vec and Random Forests also consistently improves performance over the base setting. However, the differences are statistically significant only for micro-averaged recall. 

This result is noteworthy as LINE is the embedding that has the lowest complexity of all evaluated methods which leads us to believe that the first and second order relations in multi-label networks are more important to classification quality than higher-order relationships or mesoscopic structure. We answer the question positively, and the variant of choice should be LNEMLC with LINE and Random Forest.

\subsection{How does LNEMLC perform in comparison to state of the art multi-label methods?}

\begin{figure}[htb]
    \centering
    \includegraphics[width=\columnwidth]{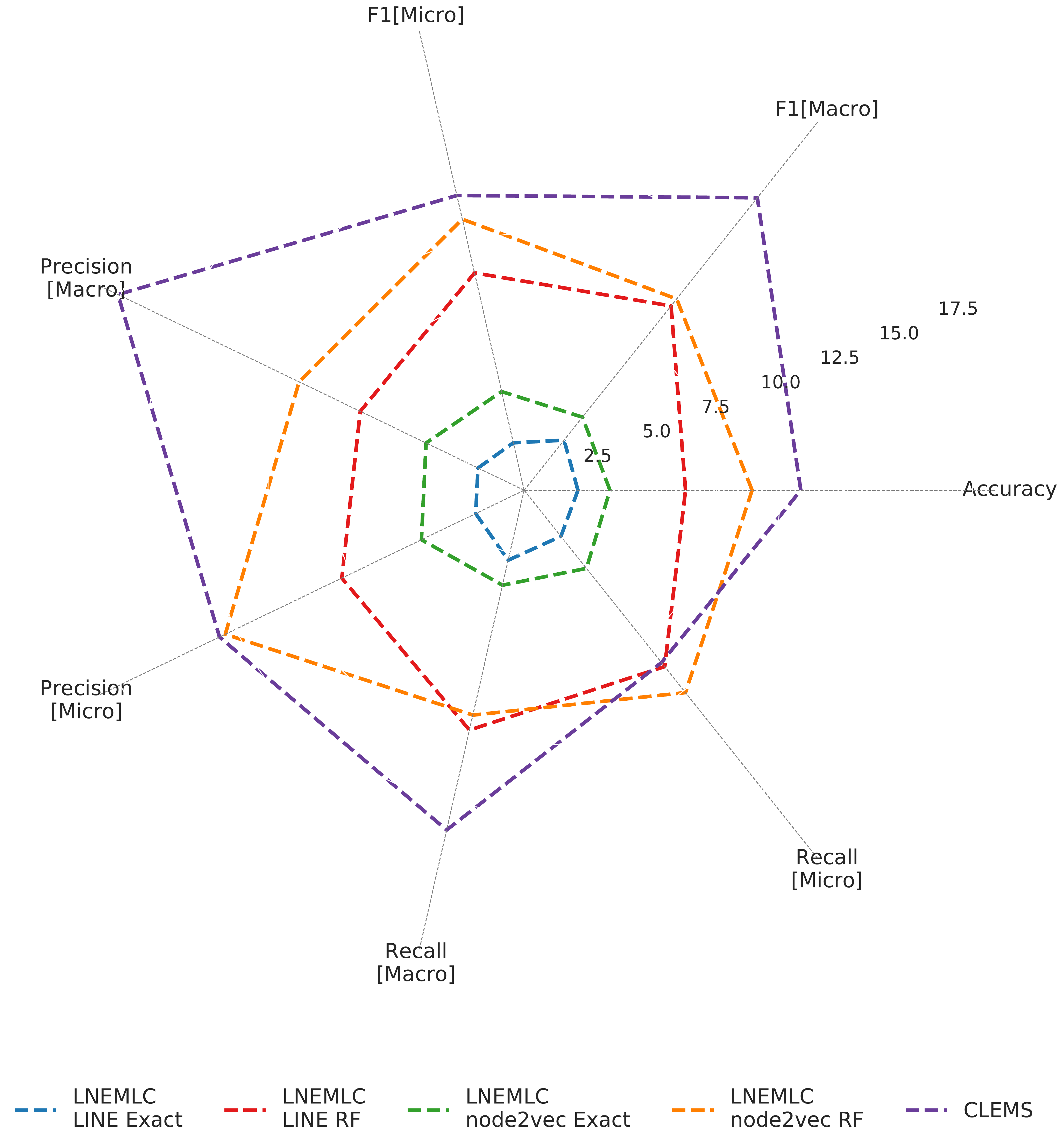}
    \caption{Mean ranks of embedding based methods in state of the art comparison among with respect to evaluated measures.}
    \label{fig:best.comparisons}
\end{figure}

Is the average 30 percentage point improvement potential of LNEMLC enough to rank higher than established multi-label approaches? Is the advantage of sustainable in the real-world scenario when regressors need to be deployed? We find out by comparing LNEMLC with LINE, and node2vec as the embedding function $\eta$ and both exact and Random Forest regressed embeddings. Results for accuracy are presented in Table \ref{tab:c.acc} while rank comparison between methods is presented in Figure \ref{fig:best.comparisons}.

\begin{table}[htb]
    \centerfloat
        \resizebox{\columnwidth}{!}{% use resizebox with textwidth
\begin{tabular}{lrrrrrrrrrr|r}
\toprule
{} & \rot{90}{bibtex} & \rot{90}{Corel5k} & \rot{90}{delicious} &  \rot{90}{emotions} &  \rot{90}{enron} & \rot{90}{mediamill} & \rot{90}{medical} &  \rot{90}{scene} & \rot{90}{tmc2007\_500} &  \rot{90}{yeast} &   \rot{90}{Mean} \\
\midrule
\textbf{LNEMLC LINE Exact}     &  0.378 &   0.394 &     0.013 &     0.223 &  0.299 &     0.801 &   0.772 &  0.974 &   0.863 &  0.944 &   \textbf{2.10} \\
\textbf{LNEMLC node2vec Exact} &  0.578 &    0.42 &     0.014 &     0.134 &  0.318 &     0.794 &   0.791 &  0.722 &   0.772 &  0.847 &   \textbf{3.35} \\
\textbf{LNEMLC LINE RF}        &  0.159 &    0.01 &     0.008 &     0.163 &  0.161 &     0.112 &    0.65 &  0.694 &   0.699 &  0.218 &   \textbf{6.30} \\
\textbf{LNEMLC node2vec RF}    &  0.181 &   0.012 &     0.007 &     0.134 &  0.149 &     0.081 &   0.583 &  0.630 &   0.625 &  0.195 &   \textbf{8.90} \\
HOMER          &  0.165 &   0.002 &     0.001 &     0.163 &  0.145 &     0.053 &    0.61 &  0.661 &   0.765 &  0.213 &   9.30 \\
RF-PCT         &  0.098 &       0 &     0.007 &     0.307 &  0.131 &     0.122 &   0.538 &  0.518 &   0.816 &  0.152 &   9.45 \\
NMF Exact      &  0.083 &     DNF &      0.01 &     0.139 &    DNF &     0.766 &     DNF &  0.866 &   0.515 &  0.927 &  10.15 \\
CC             &  0.202 &       0 &     0.006 &     0.124 &      0 &      0.08 &       0 &  0.685 &   0.787 &  0.239 &  10.45 \\
ECC            &  0.109 &   0.001 &       DNF &     0.168 &  0.131 &     0.065 &   0.526 &  0.665 &   0.608 &  0.215 &  10.50 \\
CLEMS          &  0.178 &     DNF &       DNF &     0.252 &  0.161 &       DNF &   0.664 &  0.666 &     DNF &  0.192 &  10.80 \\
BR             &  0.194 &       0 &     0.004 &     0.129 &  0.149 &      0.08 &       0 &  0.639 &   0.772 &  0.190 &  10.85 \\
RAkEL          &    DNF &       0 &       DNF &     0.208 &  0.136 &      0.06 &   0.607 &  0.694 &   0.734 &  0.201 &  10.95 \\
QWML           &  0.186 &   0.012 &       DNF &     0.149 &  0.097 &     0.044 &    0.48 &  0.630 &   0.768 &  0.192 &  11.10 \\
CLR            &  0.183 &    0.01 &       DNF &     0.144 &  0.117 &     0.044 &   0.486 &  0.633 &   0.767 &  0.195 &  11.20 \\
RFML-C4.5      &  0.011 &   0.008 &     0.018 &     0.272 &  0.124 &     0.104 &   0.216 &  0.372 &   0.421 &  0.129 &  11.40 \\
ML-C4.5        &  0.095 &       0 &     0.001 &     0.277 &   0.14 &     0.049 &   0.646 &  0.533 &   0.078 &  0.158 &  12.15 \\
NMF RF         &   0.07 &     DNF &     0.006 &     0.139 &    DNF &     0.081 &     DNF &  0.658 &   0.456 &  0.257 &  12.35 \\
MLkNN        &  0.056 &       0 &     0.003 &     0.084 &  0.062 &      0.11 &   0.462 &  0.573 &   0.305 &  0.159 &  14.05 \\
PCT            &  0.004 &       0 &     0.001 &     0.223 &  0.002 &     0.065 &   0.177 &  0.509 &   0.215 &  0.152 &  14.65 \\
\bottomrule
\end{tabular}
}
\caption{Comparison of LNEMLC, CLEMS and methods from Madjarov's comparison performance regarding accuracy on evaluated data sets. Algorithms that failed to finish are marked as DNF. Mean ranks treat algorithms that failed to finish as ex aequo last.}
\label{tab:c.acc}
\end{table}

Table \ref{tab:c.acc} shows us that LNEMLC with both LINE and node2vec exact embeddings ranks first among available methods in accuracy. LNEMLC has a higher potential to improve generalization possibilities of joint label distributions than the compared methods. With embeddings regressed using Random Forests LNEMLC with both LINE and node2vec rank higher than the rest of well-established multi-label methods. CLEMS ranks high on several data sets but fails to finish on larger data sets due to considerable complexity, while LNEMLC finishes on all data sets maintaining the ability to improve label combination assignment prediction.

LNEMLC also achieves a remarkable result of improving MLkNN's average rank by eight places, without parameter estimation for the base method, while MLkNN's score from Madjarov's study is the best one achieved after extensive parameter estimation. We see that when generalizing the joint label distribution is essential, using LNEMLC with network embeddings is a better idea then performing parameter estimation in case of the nearest-neighbour classifier.

When just the multi-label embeddings are compared, as in Figure \ref{fig:best.comparisons}, LNEMLC on average ranks better in 5 out of 6 main label-based measures from Madjarov's study, and when LINE embeddings are regressed with Random Forests it performs on par with CLEMS in the last measure (macro-averaged recall). All of this is achieved in a fraction of time needed by CLEMS because LNEMLC trains one regressor for the entire embedding space instead of one regressor for each of the embeddings dimensions. We thus answer the question no. three positively recommending LNEMLC with Random Forest regressed LINE embeddings for practical usage.

\subsection{Is there a single combination of method's parameters and configuration that works satisfactorily across data domains?}

\begin{table}[htb]
    \centering
    \resizebox{\columnwidth}{!}{\begin{tabular}{llllrrrrrrrr}
    \toprule
             &   &     &     &   \rot{90}{Accuracy} &  \rot{90}{F1[Macro]} &  \rot{90}{F1[Micro} &  \rot{90}{Precision[Macro]} &  \rot{90}{Precision[Micro]} &  \rot{90}{Recall[Macro]} &  \rot{90}{Recall[Micro]} &   \rot{90}{Mean} \\
    NV & O & $\xi$ & $d$ &       &        &        &        &        &        &        &        \\
    \midrule
    U & 1+2 & A & 1024 &   9.0 &    8.0 &    8.0 &    8.0 &    6.0 &    8.0 &    9.0 &   8.00 \\
             &   &     & 512 &   7.0 &    7.0 &    9.0 &    9.0 &    2.5 &    9.0 &    8.0 &   7.36 \\
             &   &     & $\pm5m$ &   3.0 &    6.0 &    6.0 &    7.0 &    8.0 &    7.0 &    5.0 &   6.00 \\
             &   &     & $\pm6m$ &   4.0 &    1.5 &    7.0 &    2.0 &    7.0 &    3.5 &    7.0 &   4.57 \\
             &   &     & $\pm5l$ &   1.0 &    4.5 &    1.0 &    5.5 &    2.5 &    1.0 &    4.0 &   2.79 \\
    W & 1+2 & A & 4096 &   5.5 &    1.5 &    4.0 &    1.0 &    9.0 &    3.5 &    2.5 &   3.86 \\
             &   &     & $\pm5m$ &   8.0 &    9.0 &    5.0 &    5.5 &    5.0 &    6.0 &    6.0 &   6.36 \\
             &   &     & $\pm6m$ &   5.5 &    3.0 &    2.5 &    3.5 &    1.0 &    5.0 &    1.0 &   3.07 \\
             &   &     & $\pm3l$ &  10.0 &   10.0 &   10.0 &   10.0 &   10.0 &   10.0 &   10.0 &  10.00 \\
             &   &     & $\pm5l$ &   2.0 &    4.5 &    2.5 &    3.5 &    4.0 &    2.0 &    2.5 &   3.00 \\
    \bottomrule
    \end{tabular}}
    \caption{Ranking of best performing parameter configurations for LNEMLC with random forest regressed LINE embeddings. NV is the network variant unweighted (U) or weighted (W), the order is the proximity order used for LINE embeddings 1+2 means both first and second order used together, $\xi$ is the aggregation function and $d$ are the numbers of dimensions.}
    \label{tab:parameter.variant.selection}
\end{table}

One of the problems with multi-label embeddings is its dependence on multiple parameters.  Estimating them is not an untractable burden due to the low complexity of linear embeddings, however having estimated them, we show which have the most potential. We check whether there exists a single combination, a rule of thumb so to speak, which when used, allows the method to achieve decent results. Many embedding methods state their preferred dimension sizes, especially among domain-constrained classical representation learning. 

Altogether we evaluate $2 \times 5 \times 11 \times 3 \times 3 = 990$ parameter configurations, but seeing how embeddings and regressors perform we concentrate on selecting the parameter configuration for LNEMLC with Random Forest regressed LINE embeddings. These parameters include label network variant, label vector aggregation method, the order of label relationships to take into account and dimensions size. 

We compare dimension size grouped by the concrete number: $d=...,128, 256...$, or their proportion to the label or feature count. For each data set, we take data points from our parameter estimation experimental scenario and treat each group of parameter values as a separate observation. We compare them in each measure separately. All results in the data set are then normalized by the best performance in a given measure. We consider only these parameter configurations for which more than half of the data points are in the top 1 percentile of all parameter configurations' in that measure. Additionally, we require that a selected method includes at least one data point that reaches the maximum performance. In Table \ref{tab:parameter.variant.selection} we look at the top performing parameter configurations and their average ranks.

We can see that the rule of thumb for selecting a well-performing LNEMLC parameter combination is to use a LINE embedding which takes into account both the first and second order relations in the label network and aggregate it for each of the samples using addition as is the standard approach in many embedding-based methods. Unweighted and weighted label networks yield similar performance. Best performing dimension sizes are usually greater than the number of labels because LINE embeddings join the proximity perspectives by glueing together two sub-embeddings. We recommend using unweighted or weighted label networks with dimension size set to the power of two closest to $5l$. If a fixed dimension is required $d=4096$ should be used.

\section{Conclusions and future work\label{sec:conclusion}}

We proposed a novel multi-label learning approach that uses state of the art network embedding approaches to incorporate label relationships into the feature matrix. Incorporating label relation information into the input space allows distance-based classification approaches, such as kNN, to perform better discrimination within extended feature space and correct the generalization quality on measures that require learning of the joint probability distribution. The newly proposed inference scheme includes using one regressor on all predicted embedding dimensions jointly and any multi-label base classifier.

We evaluated both, the most popular network embeddings such as node2vec or LINE, and also the more recent M-NMF one. We provided parameter selection insights for our method and selected best performing label network variants, dimension sizes and label vector aggregation functions. 

Experimental results achieved on benchmark multi-label data sets show that our approach has strong potential for improving multi-label classification. LNEMLC with label network embedded by exact LINE approach ranked best in all evaluated measures compared to current state of the art results. 

LNEMLC with Random Forest regressed LINE embeddings ranked well, often at the top, among state of the art methods, while having much lower complexity, training and test times than the current best multi-label embedding CLEMS. We also provide a well-performing proposition of parameters that can be used as defaults with no need to estimate them.

The proposed method  yielded  statistically  significant  improvements over the kNN baseline classifier extensively used in embedding methods and was shown to learn joint  conditional  distribution  under  a  new  inference  scheme for embeddings. 

Future work on LNEMLC may include evaluating other label network embeddings as the field of Network Representation Learning progresses dynamically, using more complicated embedding regressors to better harness the potential shown for the method and trying out other base classifiers than nearest neighbors. It would be also interesting to consider the label network embedding and base classifier parameter learning performed under single joint cost function as well as projecting the whole learning and inference scheme to other problems.

To invite scholars to work on improving LNEMLC and evaluate how their network embeddings apply to label networks we provide a functional implementation of LNEMLC in the 0.2.0 release of our open source scikit-multilearn library\footnote{\url{http://scikit.ml}}.

\section*{Acknowledgment}
%Nitesh's project
%Piotr's project
The project was partially supported by The National Science Centre, Poland the research project no. 2016/23/B/ST6/01735, 2016/21/N/ST6/02382 and 2016/21/D/ST6/02948, by the European Union’s Horizon 2020 research and innovation programme under the Marie Skłodowska-Curie grant agreement No 691152 (RENOIR); the Polish Ministry of Science and Higher Education fund for supporting internationally co-financed projects in 2016-2019 (agreement no. 3628/H2020/2016/2) as well as by the Faculty of Computer Science and Management, Wrocław University of Science and Technology statutory funds.

\ifCLASSOPTIONcaptionsoff
  \newpage
\fi



\bibliographystyle{IEEEtran}

\bibliography{mybibfile}


\begin{IEEEbiography}[{\includegraphics[width=1in,height=1.25in,clip,keepaspectratio]{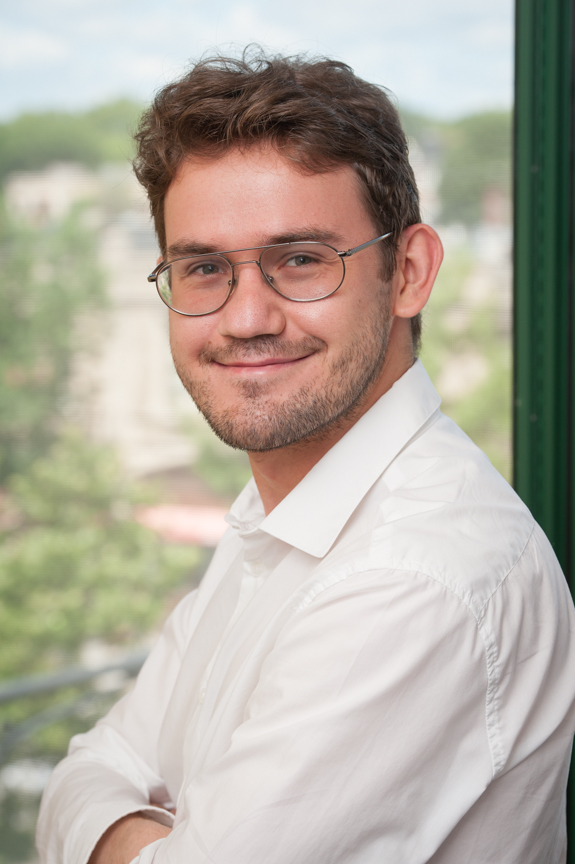}}]{Piotr Szymański}
received his M.Sc. in computer science from Wrocław University of Technology, Poland, in 2009. He had pursued a mathematics PhD 2009-2013, switching to a computer science PhD programme in 2013. He worked at the Hasso Plattner Institute in Potsdam and studied at Stanford University as a Top 500 Innovators fellow with a scholarship funded by the Polish Ministry of Science and Higher Education.He is an urban activist in Wrocław and the chairman of the Society for Beautification of the City of Wrocław; and the main author of the scikit-multilearn library.
\end{IEEEbiography}

\begin{IEEEbiography}[{\includegraphics[width=1in,height=1.25in,clip,keepaspectratio]{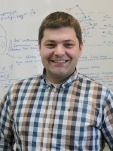}}]{Tomasz Kajdanowicz}
received his M.Sc. and Ph.D. degrees in computer science with honours, both from Wroclaw University of Technology, Poland, in 2008 and 2012, respectively. His PhD was given The Best Polish Ph.D. Disertation Award in 2012/2013, European Association for Artificial Intelligence – Polish Chapter, 2014. Recently, he serves as an assistant professor of Wroclaw University of Science and Technology at the Department of Computational Intelligence, Poland.
\end{IEEEbiography}

\begin{IEEEbiography}[{\includegraphics[width=1in,height=1.25in,clip,keepaspectratio]{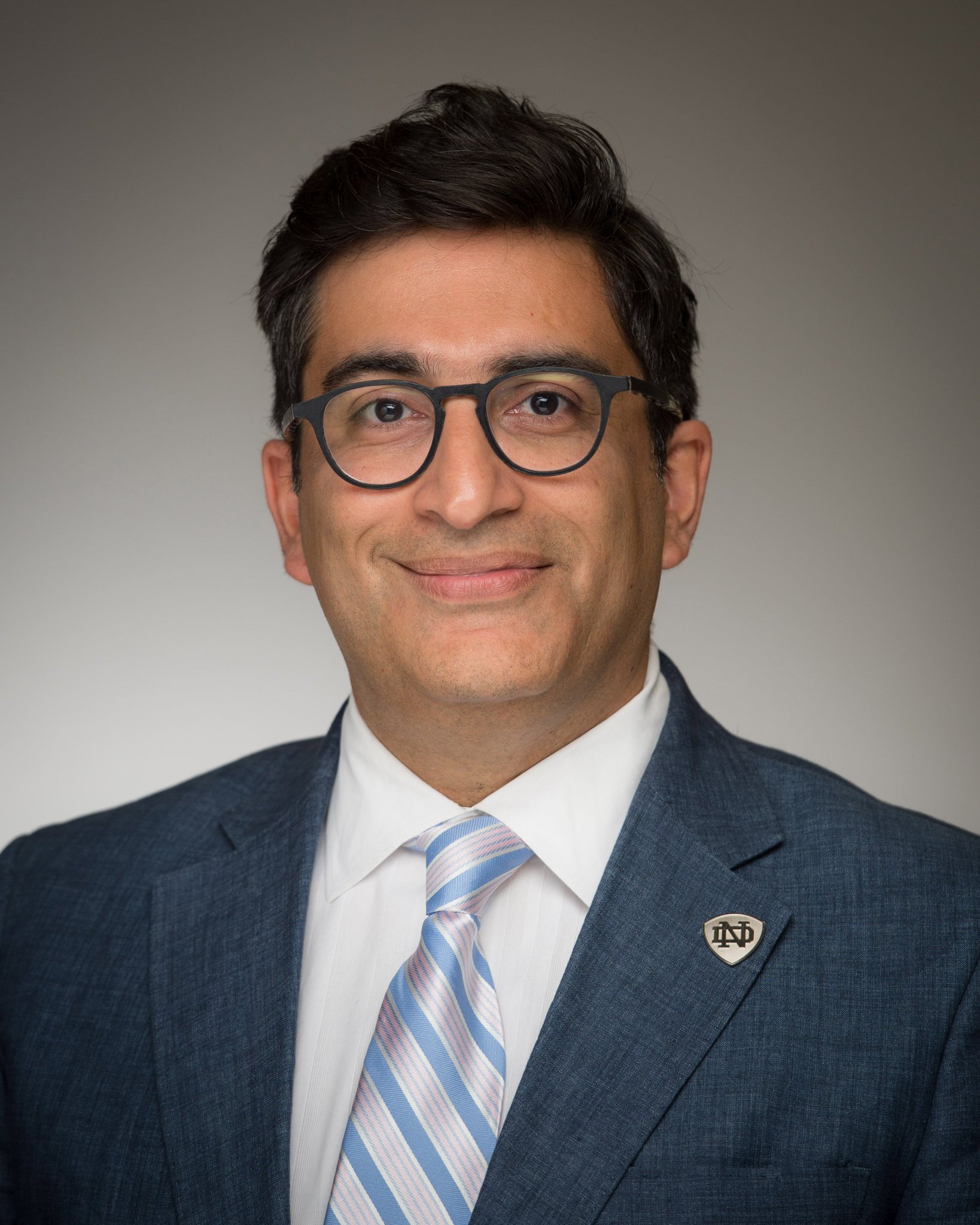}}]{Nitesh V. Chawla}
Nitesh V. Chawla is the Frank M. Freimann Professor of Computer Science and Engineering at University of Notre Dame. He is the director of the Notre Dame Interdisciplinary Center for Network Science (iCeNSA). He is the recipient of the 2015 IEEE CIS Outstanding Early Career Award for 2015, the IBM Watson Faculty Award in 2012, and the IBM Big Data and Analytics Faculty Award in 2013, the National Academy of Engineering
New Faculty Fellowship, the Rodney Ganey Award in 2014 and Michiana 40 Under 40 in 2013. He has also received and nominated for a number of best
paper awards. He serves on editorial boards of a number of journals
and organization/program committees of top-tier conferences. He is also
the director of ND-GAIN Index, Fellow of the Reilly Center for Science,
Technology, and Values, the Institute of Asia and Asian Studies, and the
Kroc Institute for International Peace Studies at Notre Dame.
\end{IEEEbiography}


%\vfill


\end{document}